\documentclass{article}
\usepackage{spconf,amsmath,graphicx}
\usepackage{amsfonts}
\usepackage{xcolor}
\usepackage{booktabs}
\usepackage{appendix}
\usepackage{newfloat}
\usepackage{subfig}
\usepackage{hyperref}
\usepackage{algorithm}
\usepackage[noend]{algpseudocode}
\DeclareMathOperator*{\argmax}{arg\,max}
\DeclareMathOperator*{\argmin}{arg\,min}

\newcommand\hw[1]{{\color{black}{#1}}}
\title{Training set cleansing of backdoor poisoning by self-supervised representation learning}
\name{
\begin{tabular}{@{}c@{}}
Hang Wan$\mbox{g}^{1,2,*}$, Sahar Karam$\mbox{i}^{1, *}$, Ousmane Di$\mbox{a}^1$, Hippolyt Ritte$\mbox{r}^1$, Ehsan Emamjomeh-Zade$\mbox{h}^1$,\\ Jiahui Che$\mbox{n}^1$,  Zhen Xian$\mbox{g}^2$, David J. Mille$\mbox{r}^{2}$,
George Kesidi$\mbox{s}^{2}$
\end{tabular}
}
\vspace{-0.2in}
\address{
\begin{tabular}{ccc}
$\mbox{ }^1$Meta
$\mbox{ }^2$Pennsylvania State University
\end{tabular}
}

\begin{document}
%\ninept
%
\maketitle
\def\thefootnote{*}\footnotetext{Equal contribution, corresponding to Sahar Karami (sahark@meta.com)}\def\thefootnote{\arabic{footnote}}
\begin{abstract}
A backdoor or Trojan attack is an important type of data poisoning attack against deep neural network (DNN) classifiers, wherein the training dataset is poisoned with a small number of samples that each possess the backdoor pattern (usually a pattern that is either imperceptible or innocuous) and which are mislabeled to the attacker's target class. When trained on a backdoor-poisoned dataset, a DNN behaves normally on most benign test samples but makes incorrect predictions to the target class when the test sample has the backdoor pattern incorporated (i.e., contains a backdoor trigger). Here we focus on image classification tasks and show that supervised training may 
%DJM -- the term "overfit" is not explained in the next sentence...
% no idea what is meant by that... and it seems an insignificant statement
% anyway -- of course the backdoors are successful -- this is well-known
% "also "strong correlation" is a vague term as well..
build stronger association between the backdoor pattern and the associated target class than that between normal features and the true class of origin.  By contrast, self-supervised representation learning ignores the labels of samples and learns a feature embedding based on images' semantic content. 
%We thus propose to use unsupervised representation learning to avoid emphasising backdoor-poisoned training samples and learn a similar feature embedding for samples of the same class.
Using a feature embedding found by self-supervised representation learning, 
a data cleansing method, which combines sample filtering and re-labeling, is developed.  Experiments on 
CIFAR-10 benchmark datasets show that our method achieves state-of-the-art performance in mitigating backdoor attacks. 
\end{abstract}

\begin{keywords}
Backdoor; contrastive learning; data cleansing
\end{keywords}
\vspace{-0.25in}
\section{Introduction}
\label{sec:intro}
\vspace{-0.1in}

It has been shown that Deep Neural Networks (DNNs) are vulnerable to backdoor attacks (Trojans) \cite{BadNet}.  Such an attack is launched by poisoning a small batch of training samples from one or more source classes
chosen by the attacker. Training samples are poisoned by embedding innocuous or imperceptible backdoor patterns into the samples and changing their labels to a target class of the attack. For a successful attack, a DNN classifier trained on the poisoned dataset: i) will have good accuracy on clean test samples (without backdoor patterns incorporated); ii) but will classify test samples that come from a source class of the attack, but with the backdoor pattern incorporated (i.e., backdoor-triggered), to the target class. 
%A backdoor attack can either happen when there is an attack that has access to part of the training set, or happen naturally due to some unreasonable labeling behaviors. 
Backdoor attacks may be relatively easily achieved in practice because of an insecure training out-sourcing process,
through which both a vast training dataset is created and deep learning itself is conducted.
Thus, devising realistic defenses against backdoor poisoning is an important research area.
In this paper, we consider defenses that operate after the training dataset is formed
but before the training process. The aim is to cleanse the training dataset prior to
training of the classifier.

%DJM -- above reference ??

%In our work, we consider a more challenging task of training-dataset cleaning which is deployed before fully training the DNN classifier.
%DJM -- again  "strong correlation" in the sentence below is a vague term - correlation
% between what and what ??

%DJM - the real inspiration is that one can largely avoid the effects of
% backdoor poisoning if we (temporarily) discard the labels, and only learn
% an unsupervised representation of the input pattern...

%DJM -- also may need to state some assumption about the attack -- if the 
% attacker chooses some poisoned samples to be very close to each other,
% it could defeat the KNN strategy...

We observe that, with supervised training on the backdoor-attacked dataset, a DNN model learns stronger ``affinity'' between the backdoor pattern and the target class than that between normal features and the true class of origin. This strong affinity is enabled (despite the backdoor pattern typically being small in magnitude) by mislabeling
the poisoned samples to the target class.
%(Indeed, the poisoned DNN is trained so that the activations due to the backdoor pattern can ``overcome" those of the salient source-class discriminative features when a backdoor trigger is input \cite{UnivBD}.) 
However, self-supervised contrastive learning,
does {\it not} make use of supervising class labels;
thus, it provides a way for learning from the training set
{\it without} learning the backdoor mapping.  

\begin{figure}[t!]
\vspace{-0.2in}
  \centerline{\includegraphics[width=8cm]{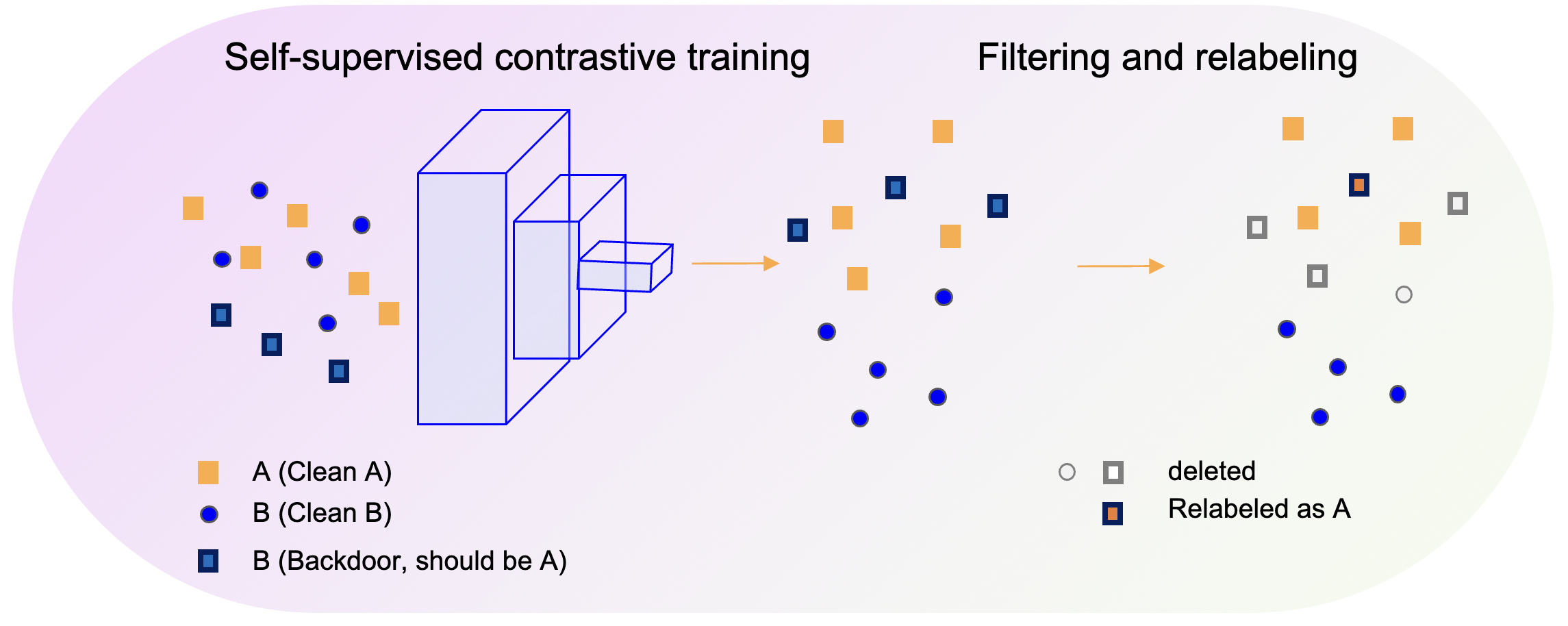}}
\caption{Overview of the data cleansing method.
}
\label{fig:overview}
\vspace{-0.25in}
\end{figure}

Based on this observation, a training set cleansing method is proposed. Using the training set $\mathcal{D}$, we first learn a feature representation using self-supervised contrastive loss. We hypothesize that, since the backdoor pattern is small in magnitude, self-supervised training will not emphasize the features of the backdoor pattern contained in the poisoned samples. Working in the learned feature embedding space, we then propose two methods (kNN-based and Energy based) to detect and filter out samples whose 
predicted class is not in agreement with the labeled class. 
We then relabel detected samples
to
their predicted class (for use in subsequent classifier training) if the prediction is made 
``with high confidence''. 
%We find experimentally that most of the backdoor examples are filtered out with nearly 90\% clean samples preserved, and the remaining backdoor samples do not have a significant impact on the DNN training since there will be samples with backdoor patterns relabeled to classes other than the target class.
%DJM -- what about false positives ?? you are not mentioning these...
%To the best of our knowledge, we are the first work that proposes a way to neutralize the backdoor sample leaking when doing training set cleaning. 
An overview of our method is shown in Fig. \ref{fig:overview}. 
\hw{Unlike many existing backdoor defenses,
Our method requires neither a small clean dataset available to the defender, nor
a reverse-engineered backdoor pattern (if present), nor a fully trained DNN classifier on the (possibly poisoned) training dataset.  Also, ours is the first work to address the problem of backdoor samples evading (``leaking through'') a rejection filter -- we propose a relabeling method to effectively neutralize this effect.}
%Also need to say that unlike our previous COSE paper we do not rely on
% knowledge of the backdoor embedding function...
%Though we focus on image classification, our method is easily generalized to other domains. 
A complete version of our paper with Appendix is \href{https://arxiv.org/pdf/2210.10272.pdf}{online} available.

\vspace{-0.1in}
\section{Threat model and related works}
\label{sec:threaten}
\vspace{-0.1in}

Consider a  clean dataset ${\mathcal D} = \{(x_i, y_i)~|~ i = 1 ... N\}$, where: $x_i \in {\mathbb R}^{X \times H \times W}$ is the $i^{\rm th}$ image in the dataset with $X$, $H$ and $W$ respectively the number of image channels, height, and width;  $y_i \in \{1,2,...,C\}$ is the corresponding class label, with the number of classes $C>1$. Backdoor attacks poison a dataset by: i) choosing an attack target class $t$, and then obtaining a subset (of size $M$) of images from classes other than $t$: ${\mathcal D}_s = \{(x_j, y_j)| i = 1 ... M, y_j \neq t\}$, ${\mathcal D}_s \subset {\mathcal D}$, and $M \ll N$; ii) the backdoor pattern is then incorporated into each sample in ${\mathcal D}_s$ using the attacker's backdoor embedding function $g: {\mathbb R}^{X \times H \times W} \xrightarrow{} {\mathbb R}^{X \times H \times W}$; iii) the label for each poisoned sample is then changed to the target class: ${\mathcal D}_p = \{(g(x), t)| x\in {\mathcal D}_s\}$; iv) finally the poisoned dataset is formed by putting the attacked images back into the training set: $\Bar{{\mathcal D}} = ({\mathcal D} \backslash {\mathcal D}_s) \cup {\mathcal D}_p$. If the attack is successful, the victim model $f: {\mathbb R}^{X \times H \times W} \rightarrow \{1,2,...,C\}$, when trained on the poisoned dataset, will have normal (good) classification accuracy on clean (backdoor-free) test samples, but will classify
most backdoor-triggered test samples to the target class of the attack. In the image domain, backdoor patterns could, e.g., be: i) a small patch that replaces the original pixels of an image \cite{BadNet, NC, MAMF}; ii) a perturbation {\em added} to some pixels of an image \cite{Targeted, Haoti, Post-TNNLS}; or iii) a ``blended'' patch attack \cite{Targeted}. 
%The detailed explanation and some examples of backdoor images can be found in Appendix XXX. 

On the other hand, the defender aims to obtain a classifier with good classification accuracy on clean test samples and which correctly classifies test samples with the backdoor pattern.
Defenses against backdoors that are deployed post-training aim to detect whether a DNN model is a backdoor victim \cite{NC, Tabor, ABS, META, Post-TNNLS, UnivBD} and, further, to mitigate the attack if a detection is declared \cite{NC, NAD, FP}.
%For backdoor detection, one common practice is to reverse-engineer backdoor patterns  for each putative target class. An detection is declared if an abnormally small pattern (when embedded into images) can induce the mis-classification of all the images \cite{}. Meta learning methods assume the availability of some shadow clean and attacked models, a binary classifier can be trained based on the features extracted from the shadow models and used for detection \cite{META}. \cite{UnivBD} show that backdoor attack model has abnormally large {\it maximum margin} which can be used for detection. 
%Mitigation methods aim to make the DNN `unlearn' the backdoor triggers such that a attacked test sample will be classified into it's original class. It is shown that backdoor can be mitigated by pruning \cite{FP, },  fine-tuning \cite{NC, NAD}, or activation bounding \cite{UnivBD}. 
Most post-training defenses 
%do not assume access to the potentially poisoned 
%training set,
%but 
require a relatively small clean dataset (distributed as the clean training set), with their performance generally sensitive to the number of available clean samples \cite{FP, NC, Tabor, UnivBD}. 
%Also, some post-training defenses attempt to reverse-engineer the backdoor pattern, e.g.,
%\cite{NC,Post-TNNLS,MAMF}, and thus assume knowledge of the backdoor embedding function (which may not be available in practice).
%Test-time defenses aim to detect a backdoor trigger 
%\cite{NC,B3D,STRIP,InFlight}; the first three requiring careful choice of hyperparameters 
%for good performance, and the latter leverages reverse-engineering of the backdoor pattern.
In this paper, alternatively, we aim to cleanse the training set prior to deep learning. Related work on training set cleansing includes \cite{AC,SS,CI,CandS}.
%\cite{UnivBD}
All of these methods rely on embedded feature representations
of a classifier fully trained on the possibly
poisoned training set (\cite{SS} suggests that an auto-encoder could
be used instead). \cite{SS,AC} use a 2-component clustering
approach to separate backdoor-poisoned samples from clean samples
(\cite{SS} uses a singular-value decomposition while \cite{AC} uses
a simple 2-means clustering), 
while \cite{CI} uses a Gaussian mixture model whose number of
components is chosen based on BIC \cite{Schwarz}. Instead of clustering,
\cite{CandS} employs a reverse-engineered backdoor pattern estimated
using a small clean dataset.
DBD \cite{Decouple_yiming} 
builds a classifier based on an encoder learned via self-supervised contrastive loss; then the classifier is fine-tuned. In each iteration some samples are identified as ``low credible'' samples by the classifier, with their labels removed; the classifier is then updated based on the processed dataset in a semi-supervised manner.

\vspace{-0.25in}
\section{Methodology}
\vspace{-0.1in}

\subsection{Vulnerability of supervised training}
\vspace{-0.05in}
We now illustrate the vulnerability of supervised training by analysis of a simple linear model trained on a poisoned dataset, considering the case where all classes other than the target are (poisoned) source classes. The victim classifier forms a linear discriminant function for each class  $s$,  i.e., the inner product $f_s(x) = x \cdot w_x $, 
where $w_s \in{\mathbb R}^{X \times H\times W}$ is the vector of model weights corresponding to class $s$.  
%(Alternately, one can use a first-order Taylor approximation of a nonlinear discriminant function, which
%is justified because a covert backdoor pattern will have small magnitude.)
%f(x+Deltax) = f(x) + nabla f_s (x) cdot Deltax
Assume that, after supervised training, each training sample is classified correctly with confidence
at least $\tau>0$ as measured by the margin:
\vspace{-0.05in}
\begin{equation}
\label{eq:margin}
    f_{y_i}(x_i)  - \max_{c \neq y_i}f_c(x_i) \geq \tau, ~~\forall (x_i, y_i) \in \Bar{{\mathcal D}}.
    \vspace{-0.1in}
\end{equation}

Assuming that the backdoor pattern $\Delta x$ is additively incorporated, given an attack sample based on clean $x_s$ originally from source-class $s\not=t$, Eq. \eqref{eq:margin} implies
\vspace{-0.05in}
\begin{equation}
\label{eq:attack_margin}
    w_t \cdot (x_s + \Delta x) - w_s \cdot (x_s + \Delta x ) \geq \tau.
    \vspace{-0.05in}
\end{equation}
If $x_s$ is also classified to $s$ with margin $\tau$, then 
\vspace{-0.07in}
\begin{equation}
\label{eq:clean_margin}
    w_s \cdot x_s  - w_t \cdot x_s \geq \tau.
    \vspace{-0.07in}
\end{equation}
Adding \eqref{eq:margin} and \eqref{eq:clean_margin} gives\vspace{-0.05in}
\begin{equation}
\label{eq:perturb_margin}
f_t(\Delta x) - f_s(\Delta x) = (w_t - w_s) \cdot \Delta x \geq  2\tau. 
\vspace{-0.05in}
\end{equation}
This loosely suggests that, after training with a poisoned training dataset, 
the model has {\it stronger} ``affinity"
between the target class and the backdoor pattern \eqref{eq:perturb_margin} than between the source class and the class-discriminative features of clean source-class samples \eqref{eq:clean_margin}. This phenomenon is experimentally verified when the model is a DNN, as shown in Apdx. \ref{apdx:dnn}. 

%This phenomenon holds for covert backdoor patterns, which are either imperceptible or innocuous to a human observer (thus small in magnitude), so that the class-discriminative features of the clean source-class samples are still strongly expressed in the poisoned training samples created from them.
%DJM-new -- crucial connecting sentence
However, these strong affinities are only made possible by the mislabeling
of the backdoor-poisoned samples.
Given that usually the perturbation $\Delta x$ is small, backdoor attacked images differ minutely from the original (clean) images. 
%DJM-new -- I changed the sentence below...
Thus if a model is trained {\it in a self-supervised manner}, 
without making use of the class labels,
the feature representations of $x$ and $x + \Delta x$ should be quite similar/highly proximal. Thus, in the model's representation space, poisoned samples 
may ``stand out''  as outliers in that their labels may disagree with the labels of samples in close proximity to them. 
This is the basic idea behind the cleansing method we now describe.

\vspace{-0.15in}
\subsection{Self-supervised contrastive learning}
\vspace{-0.05in}
\label{sec:SimCLR}

%\gk{GK: note that I added $a(i)$ to the denominator in the display below and
%I clarified its definition - formerly you called it $j(i)$.}

SimCLR \cite{SimCLR,SupCon} is a self-supervised training method to learn a feature representation for images based on their semantic content. In SimCLR, in each mini-batch, $K$ samples are randomly selected from the training dataset, and each selected sample $x_k$ is augmented to form two versions, resulting in $2K$ augmented samples. Augmented samples are then fed into the feature representation model, which is an encoder $E(\cdot)$ followed by a linear projector $L(\cdot)$, with the feature vector $z$ extracted from the last layer: $z = L(E(x))$. 
For simplicity we will refer to $L(E(\cdot))$ as the ``encoder'' hereon.
The encoder is trained to minimize the following objective function:
%DJM-new: you can't call L() both the linear projector AND the loss
% function !  I changed the loss to C below...
\vspace{-0.05in}
\begin{equation}
\label{eq:SimCLR}
C = -\frac{1}{2K}\sum_{i=1}^{2K}\log \frac{\exp{(z_i \cdot z_{\alpha(i)} / \tau})}{\sum_{i'=1,~i' \neq i}^{2K}\exp{(z_i \cdot z_{i'} / \tau})},
\vspace{-0.05in}
\end{equation}

where $i$ and $\alpha(i)$ are indexes of two samples augmented from the same 
training sample. 
%DJM -- what does the above mean -- alpha(alpha(i)) ? Is it necessary to
% say this ??
Consistent with minimizing (\ref{eq:SimCLR}), SimCLR trains the encoder by projecting an image and its augmentations to similar locations in the derived feature space. So, given the fact that a backdoor attack makes minor changes to a poisoned training sample while preserving the semantic content related to its source class (the label of the clean sample), it is expected that the encoder will learn to project a backdoor image into a feature space location close to clean (and augmented) images from the source class. Working in the feature representation space learned by SimCLR, a training-sample filtering and re-labeling method can thus be deployed to cleanse the training set. 

\vspace{-0.15in}
\subsection{Data filtering}
\vspace{-0.05in}
We consider two options for data filtering: k Nearest Neighbor (kNN) classifier and class-based ``energy'' score.

{\em kNN}: kNN is a widely used classification model. The class of a sample is determined by the voting of its top k nearest samples in the derived feature space. Basically we want to leverage the SimCLR representations and compare the label of a data point in the training set with the labels of nearby data points in representation space to verify the class label.  The class label of a training sample is accepted if its labeled class agrees with kNN's predicted class (based on plurality voting); otherwise it is likely to be mislabelled/attacked and will be rejected. In our experiments, $k$ is chosen to be half of the number of images from each of the classes\footnote{We found experimentally that this large choice of $k$ yields more accurate filtering than smaller choices of $k$.}.

%DJM-new -- how do you justify choosing k this large ?
%DJM -- did you use simple plurality logic ??
%DJM -- in the above, why majority logic instead of plurality logic ??
%Hang yes it should be plurality logic

{\em Energy}: Given a sample (in the embedded feature space) $z_i$,  an
energy score corresponding to class $c$ is as follows:
\begin{equation}
    S_c(z_i)=\log \frac{1}{|{\mathcal I}_c\backslash\{i\}| } \sum_{k\in {\mathcal I}_c\backslash\{i\}} \frac{\exp(z_i\cdot z_k/\tau)}{\sum_{k'=1, k'\not =i}^N \exp(z_i \cdot z_{k'}/\tau)},
\end{equation}
where ${\mathcal I} = \{i ~|~ i=1...N\}$ is the set of indices of all the samples in the training set and ${\mathcal I}_c = \{i ~|~ i\in {\mathcal I}, y_i=c\}$ is the set of indices of samples from class $c$.
A class decision can be made based on a sample's class-conditional scores: $c^*(z_i) = \argmax_c S_c(z_i)$. A training sample is accepted only if its predicted class $c^*(z_i)$ agrees with its class label.

Tab. \ref{tab:filter} shows the performance of Energy and kNN filtering methods using the ResNet-18 encoder architecture on the CIFAR-10 dataset, when 1000 samples are poisoned. Our filtering method can filter out 97\% of backdoor samples while keeping most of the clean samples. However, the leaking of even a few backdoor samples is still
problematic -- \cite{BadNet} shows that even 50 backdoor samples for the CIFAR-10 dataset can make the attack successful\footnote{More than 90\% of test images with the backdoor trigger are classified to the target class.}. So one cannot successfully defend backdoor attacks only by data filtering.
%DJM -- you could filter out the UNION of the Energy and KNN detections...
\begin{table}[ht!]
\vspace{-0.1in}
    \scriptsize
    \centering
    \begin{tabular}{ccc}\toprule
                                &  clean  &    backdoor \\\hline\hline
Energy                   &  89.14 &    2.9      \\\hline
kNN                      &  88.95 &    3.2     \\\bottomrule
    \end{tabular}
    \vspace{-0.1in}
    \caption{The percentage of clean and backdoor images remaining after filtering methods applied. %ResNet-18 encoder is trained on CIFAR-10 dataset poisoned with 1000 backdoor attacked images.}
    }
    \label{tab:filter}
    \vspace{-0.25in}
\end{table}

\vspace{-0.1in}
\subsection{Re-labeling}
\vspace{-0.1in}

From the discussion in Sec. \ref{sec:SimCLR}, SimCLR projects a backdoor image to a feature space location close to clean images from the source class. For a sample that is rejected with a certain level of confidence (e.g., for KNN, if all $k$ neighbors are labeled to class 1, but the sample is labeled to class 2), it is likely that the sample is a backdoor image and that the predicted class is the source class. 
Also, \cite{NC} indicates that a backdoor attack can be unlearned if a model is trained on images with backdoor triggers but labeled to the true class. Thus, to neutralize the influence of the backdoor images leaked in the filtering step, we first identify samples that are rejected \hw{(samples with kNN/energy's predictions do not agree with their class labels)} with a confidence threshold $T$.  Then we re-label these samples to the predicted class. For kNN, the confidence can be measured by the fraction of samples with the predicted label amongst the $K$ nearest neighbors; for the energy-based method, the maximum score (over all classes) can be used as the confidence measure. Note that $T$ is a hyper-parameter -- the performance of our relabeling method is sensitive to the threshold $T$. Care should be taken when a sample is relabeled since it inevitably induces label noise--the predictions of kNN and the energy-based method are not always reliable. So it is not a good idea to pre-define a number/ratio of samples that will be relabeled given that the number of backdoor attacked samples is not known. Alternatively, the confidence can be determined based on the confidence of clean samples (here we treat all accepted samples in the filtering step as clean samples). In practice, we set the threshold as the 80-th percentile of the confidences of accepted samples. 
%\hw{$T$ is the only hyper-parameter in our method and can be set automatically based on the conficence on the accepted samples.}

\vspace{-0.1in}
\section{Experiments}
\vspace{-0.1in}
%\gk{GK: compare your method against SS,AC,CI and TC-RED \cite{CandS} 
%by QUOTING results for these methods from \cite{CI,CandS}. DO NOT redo
%experiments for these previous methods. Mention that the results are
%from \cite{CI} or \cite{CandS} in the table-captions where they're reported.

%Also, consider a completely clean training dataset to 
%see how your method performs on false positives...}

Our experiments are mainly conducted on CIFAR-10 \cite{CIFAR10}.
%an image classification set with 50k images in the training set and 10k images in the test set; there are 10 classes and the image size is 32x32. 
%Our method is also evaluated on MNIST \cite{MNIST} and CIFAR-100 \cite{CIFAR10}, with these results found in Apdx. \ref{apdx:other}.
Two global backdoor patterns (additive \cite{Post-TNNLS} and WaNet \cite{nguyen2021wanet}) and two local patterns (BadNet\cite{BadNet}, blended\cite{Targeted}) are considered in our work. The details of those attacks can be found in Apdx. \ref{apdx:detail_of_attacks}. For all the experiments ResNet-18 is used as the model architecture. 
A ResNet-18 encoder is first trained for 1000 epochs on the given dataset using the SimCLR loss (Eq. \ref{eq:SimCLR}); then the kNN or energy based filtering and relabeling are applied to clean the dataset. 
Finally, the  classifier is trained on the cleaned dataset using Supervised contrastive loss (SupCon) \cite{SupCon}. The performance of a defense is evaluated by two measures: clean test accuracy (ACC) and attack success rate (ASR)\footnote{ASR: the ratio of test samples with the backdoor trigger that are classified to the target class. A lower ASR implies a better defense method.}. Our method is compared with three baseline methods: DBD\cite{Decouple_yiming}, Activation Clustering (AC)\cite{AC}, and Spectral Signature (SS) \cite{SS}. AC and SS train a ResNet-18 classifier on the poisoned dataset 
%DJM-new -- you don't say what DNN architecture was used for
% the classifier for AC and SS -- neither here nor below (!)
and then identify backdoor samples using the classifier's internal layer activations, by activation clustering (AC) or by checking if there are abnormally large activations (SS). Notably both AC and SS assume knowledge of the attacker's target class, and SS further assumes the number of poisoned images is known. After filtering, a new classifier is trained on the cleaned dataset using SupCon. DBD defends backdoor attacks by iteratively identifying backdoor images using a classifier (pre-trained using SimCLR at the beginning) and then fine-tuning the classifier using a semi-supervised loss (MixMatch \cite{mixmatch}), treating the backdoor images as unlabeled data and clean images as labeled data. \hw{The methods mentioned above
require careful adjusting of hyper-parameters to get good performance;
by contrast, there is only one hyper-parameter in our method (the threshold $T$
in the relabeling step), which can be automatically set based on the
confidence on accepted samples, as discussed above.}

\begin{table}[ht!]
\vspace{-0.1in}
	\scriptsize
	\begin{center}
			\begin{tabular}{ccccccccc}
				\toprule
				& \multicolumn{2}{c}{Additive} &
				\multicolumn{2}{c}{BadNet}& \multicolumn{2}{c}{blend} & \multicolumn{2}{c}{WaNet}\\
				\cmidrule(lr){2-3} \cmidrule(lr){4-5}
				\cmidrule(lr){6-7} \cmidrule(lr){8-9}
				&ASR&ACC&ASR&ACC&ASR&ACC&ASR&ACC
				\\\hline \hline
				None &99.9&94.14&98.4&94.19&99.9&94.14& 67.6 & 93.11  \\ 
				AC\cite{AC}&100& 92.75 &3.2 & 94.49&1.7 &94.50& 68.9 & 87.05\\
				SS\cite{SS}& 97.4 & 91.14 & 3.2&94.49 &49.9 &94.36 &63.5&93.04\\
			    DBD&0.7&92.57&1.5&91.49&1.4&91.89&0.7 & 91.65\\
				kNN &2.6&91.46& 4.2 & 92.37 &4.2 & 92.13& 3.6 & 92.24 \\
				Energy& 2.8 & 91.91& 4.9 & 92.12& 2.5& 92.10& 1.3 & 91.25\\
%				kNN + Energy\\ in appendix
		\bottomrule
		\end{tabular}
		\vspace{-0.05in}
		\caption{ASR (\%) and ACC (\%) for our methods and comparison methods under different attacks. \vspace{-0.25in}}
	\label{tab:performance}
	\end{center}
\end{table}

\begin{figure}[!ht]
    \vspace{-0.27in}
	\centering
	\subfloat[\centering Additive attack ]{{\includegraphics[width=4cm]{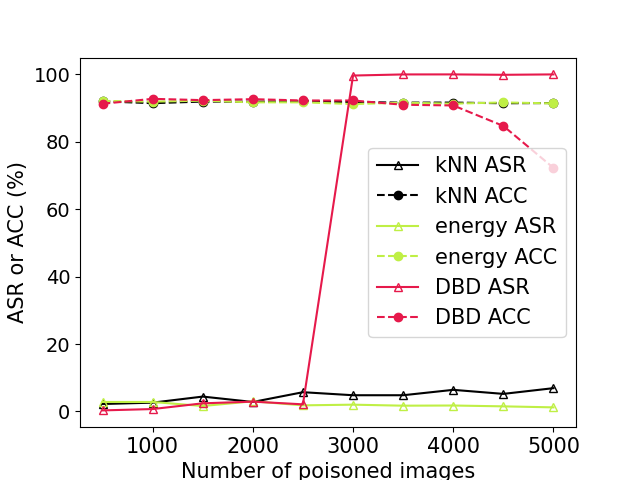} }}%
	~
	\subfloat[\centering blended ]{{\includegraphics[width=4cm]{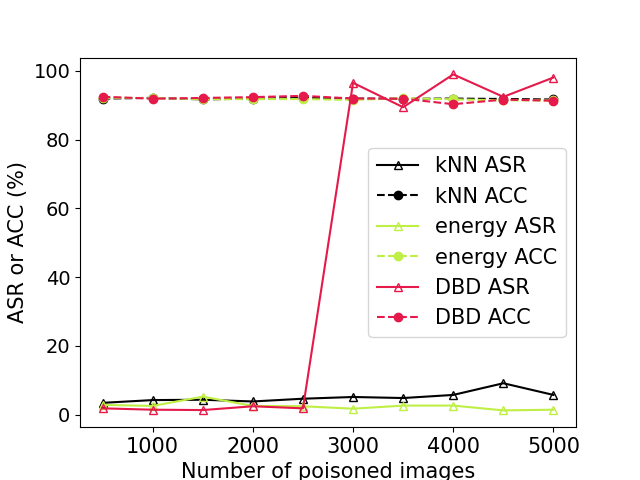} }}%
	\vspace{-0.05in}
	\caption{Performance of our methods and DBD \cite{Decouple_yiming} for different number of attack images.}
	\label{fig:different_number}
\end{figure}

{\bf Performance:}
The defense methods aim to achieve high ACC and low ASR.
Tab. \ref{tab:performance} shows the performance of different methods when the number of poisoned images is 1000 for additive, BadNet, and blend attacks and 5000 for the WaNet attack.  AC performs well for local backdoor attacks (BadNet and blend) where a backdoor image contains mostly features from the source class (original class before attack), i.e. such that the internal layer features of the classifier are different from those of clean target class images; but when the pattern is global (additive and WaNet) the classifier trained on the poisoned data can project images with backdoor patterns into locations close to those of clean target class samples; thus the performance for these attacks is poor. Similarly, SS performs well only on BadNet and can mitigate the ASR for the blended attack, but does not work well on additive and WaNet attacks. Our method and DBD give comparable performance; however Fig. \ref{fig:different_number} shows that, with an increased number of poisoned images, DBD fails (ASR close to 100\% and drop of ACC under the additive attack). 
The likely reason is the
iterative filtering of DBD -- in any iteration, if some backdoor samples are falsely identified as clean, the subsequently fine-tuned classifier may (re-)learn the backdoor pattern; thus in the next iteration 
more backdoor images will likely be falsely accepted as clean. However, our method performs only one filtering step.  Moreover, some images will be relabeled to neutralize the influence of the backdoor samples leaked in the filtering step; thus our method achieves strong robustness against backdoor poisoning even when 10\% of the training samples are poisoned. Moreover, Apdx. \ref{apdx:adaptive} 
considers an adaptive attack scenario, with the results showing that our method can mitigate the attack even when the attack assumes full knowledge of our defense method.

\begin{table}[ht!]
\vspace{-0.25in}
    \scriptsize
    \centering
    \begin{tabular}{ccccc}\hline
             &clean& backdoor&  ASR (\%)  &    ACC (\%) \\\hline\hline
baseline, without any step&100 &100&   98.4 & 94.19            \\
AE + filtering + relabeling  &22.31&5.3 & 3.7  &  51.15     \\
SimCLR + filtering           &88.95&3.2&  68.9  & 91.94         \\
SimCLR + relabeling &100 & 100& 79.8 & 94.06 \\
SimCLR + filtering + relabeling &88.95&3.2& 4.2 & 92.37\\\hline
    \end{tabular}
    \vspace{-0.05in}
    \caption{The percentage of clean (second row) and backdoor (third row) images remaining, ASR, and ACC after applying different combinations of our algorithm's components. 1000 images were poisoned by BadNet attack.\vspace{-0.15in}}
    \label{tab:ablation}
\end{table}

{\bf Ablation study:}
In our method, three main steps are deployed for data cleaning: self-supervised contrastive training (SimCLR), data filtering, and data relabeling. Here we conduct an ablation study to understand the importance of each step. For the case without SimCLR, we still need to learn an encoder for feature embedding; an auto-encoder (AE) model is used as a substitute for SimCLR. KNN is used as the filtering method. From Tab. \ref{tab:ablation}, We can see SimCLR is important for getting good embedding representations to accept most clean examples. The filtering method can reject most of the backdoor images, but if used without the relabeling, the remaining backdoor images still make the ASR high (68.9\%). Also, without filtering, relabeling some backdoor samples can only slightly mitigate the ASR (to 79.8\%). With the combination of SimCLR, filtering, and relabeling, our method achieves robustness against attacks even with a large number of poisoned images (shown in Fig. \ref{fig:different_number}).

\vspace{-0.1in}
\section{Conclusions}
\vspace{-0.05in}
In this paper, we proposed a training set cleansing method against backdoor attacks. We discussed the vulnerability of supervised trained models and thus proposed to use self-supervised learned representation embedding coupled with data filtering and relabeling. 
Experiments show that our method is robust under different types of attacks and different attack strengths.
%DJM-new -- added the next sentence
In future, this approach could be investigated on other architectures like ViT and to mitigate
non-backdoor data poisoning attacks, e.g. label flipping attacks. 
\bibliographystyle{IEEEbib}
\bibliography{reference}

\newpage

\appendix
\section{Vulnerability of supervised DNN training}
\label{apdx:dnn}

\begin{figure}[!ht]
	\centering
	\subfloat[\centering attack image ]{{\includegraphics[width=3cm]{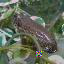} }}%
	~
	\subfloat[\centering salient map]{{\includegraphics[width=3cm]{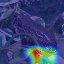} }}%
	\caption{(a): A test image with backdoor pattern. (b) the salience map produced using GradCAM on the attacked model. The test image is from Tiny-ImageNet dataset\cite{ImageNet}.}
	\label{fig:dnn}
\end{figure}

Fig. \ref{fig:dnn} shows the salience map produced by GradCAM on the poisoned classifier, for an attacked test image. The salience map shows the locations in the image that the network is focusing on when making a decision. The salience map indicates that, after supervised training, when the backdoor trigger occurs, the model will focus on the backdoor trigger and ignore the other features in an image when making a decision. This indicates that the supervised trained networks build stronger ``affinity'' between the target class and the backdoor trigger such that the backdoor trigger will dominate the DNN's decision-making even in the presence of clean (source class) features.

\section{algorithm}
The algorithm of our method is given in Algorithm \ref{alg:clean}

\begin{algorithm}[h]
	\caption{The algorithm of our training set cleansing method.}\label{alg:clean}
	\begin{algorithmic}[1]
		\State {\bf Input}: dataset ${\mathcal D} = \{(x_i, y_i)~|~ i = 1 ... N\}$.
		\State {\bf Initialization} A clean set ${\mathcal D}_{clean} = \{\}$; a bad set ${\mathcal D}_{bad} = \{\}$; and $\lambda = 80$ which is used to determine the threshold of relabeling.
		\State {\bf Self-supervised training}:
		Learn an encoder $L(E(\cdot))$ based on the given dataset ${\mathcal D}$ using the SimCLR objective (Eq. \ref{eq:SimCLR}).
		\For{$i = 1: N$} \Comment{filtering}
		\State $z_i = L(E(x_i))$
		\State $c_i = kNN(x_i)$ or $Energy(x_i)$ \Comment{$c_i$ is the class decision of the kNN or Energy method.}
		\If{$c_i == y_i$} 
		\State ${\mathcal D}_{clean}$.{\bf insert}($(x_i, y_i)$)
		\Else
		\State ${\mathcal D}_{bad}$.{\bf insert}($(x_i, y_i)$)
		\EndIf
		\EndFor
		\State ${\mathcal C} = \{ Conf_{c_i}(x_i)~|~x_i \in {\mathcal D}_{clean}\}$ \Comment{$Conf(\cdot)$ is the confidence of kNN or Energy's decision.}
		\State $T =$ {\bf percentile}$({\mathcal C}, \lambda)$ \Comment{80-th percentile of the confidences.}
        \For{$(x_j, y_j) \in {\mathcal D}_{bad}$} \Comment{relabeling}
        \If {$ Conf_{c_j}(x_j) > T $ }
        \State ${\mathcal D}_{clean}$.{\bf insert}($(x_j, c_j)$)
        \EndIf
        \EndFor
    \State {\bf Supervised training}: Train a classifier $f(\cdot)$ using SupCon\cite{SupCon} objective on ${\mathcal D}_{clean}$.
		\State {\bf Outputs}: ${\mathcal D}_{clean}$; Classifier $f(\cdot)$.
	\end{algorithmic}
\end{algorithm}

\section{Details of attacks}
\label{apdx:detail_of_attacks}

\begin{figure*}[t]
	\vspace{-0.3in}
	\centering
	\begin{minipage}[b]{.18\linewidth}
		\centering
		\centerline{\includegraphics[width=.6\linewidth]{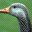}}
	    \centerline{\includegraphics[width=.6\linewidth]{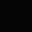}}
		\footnotesize{ ``chessboard'' (additive)}\label{fig:bd_example_A1}
	\end{minipage}
	\begin{minipage}[b]{0.18\linewidth}
		\centering
		\centerline{\includegraphics[width=.6\linewidth]{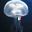}}
		\centerline{\includegraphics[width=.6\linewidth]{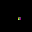}}
		\footnotesize{BadNet}\label{fig:bd_example_A2}
	\end{minipage}
	\begin{minipage}[b]{0.18\linewidth}
		\centering
		\centerline{\includegraphics[width=.6\linewidth]{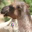}}
		\centerline{\includegraphics[width=.6\linewidth]{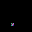}}
		\footnotesize{blend}\label{fig:bd_example_A3}
	\end{minipage}
	\begin{minipage}[b]{0.18\linewidth}
		\centering
		\centerline{\includegraphics[width=.6\linewidth]{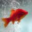}}
		\centerline{\includegraphics[width=.6\linewidth]{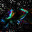}}
		\footnotesize{WaNet}\label{fig:bd_example_A4}
	\end{minipage}
	\caption{Example BPs considered in our main paper (top) and images with these BPs embedded (bottom). The images are from Tiny-ImageNet dataset \cite{ImageNet}.}
	\label{fig:bd_example}
	\vspace{-0.1in}
\end{figure*}

In the main paper, four types of backdoor attacks are used to evaluate our method: additive attack \cite{Post-TNNLS}, BadNet\cite{BadNet}, blended attack\cite{Targeted}, and WaNet\cite{nguyen2021wanet}. 
Some examples of poisoned images and backdoor patterns can be found in Fig. \ref{fig:bd_example}.

In an additive attack, a backdoor pattern embedding function is given by $g(x, v) = [x + v]_{\rm c}$, where x represents the original image,
$v$ is a small and usually imperceptible perturbation, and $[\cdot]_{\rm c}$ is a domain-dependent clipping function. The additive pattern $v$ can be either global or local. In our paper we used a global ``chessboard'' pattern, where one of two adjacent pixels are perturbed by 3/255 in all color channels.

BadNet uses patch replacement backdoor patterns embedded by $g(x, m, u) = (1-m) \odot x + m \odot u$ where $m$ is a $3\times 3$ mask and u is a $3\times 3$ noise patch. Both the location of the mask and the patch are randomly generated. 

In the blended attack, a noise patch $u$ is blended into an image using the embedding function $g(x, \alpha, m, u) = (1-\alpha \cdot m) \odot x + \alpha \cdot m \odot u$ where $m$ is a $3\times 3$. The mask is chosen to be $3\times 3$ and the patch $u$ is a $3\times 3$ noise patch. Both the location of the mask and the patch are randomly generated. The blending rate $\alpha$ is set to 0.4 in our experiments. 

WaNet is a sample-specific backdoor attack, which generates a backdoor pattern based on the image to be poisoned. In \cite{nguyen2021wanet}, the attacker is assumed to control the training process: in each training mini-batch, 10\% of the images are randomly selected and poisoned. However, our defense method is applied before classifier training, so in our work, we poisoned 10\% of the whole training set before training, which makes the ASR lower than that reported in the original WaNet paper when there is no defense.

\section{Experiments on other datasets}
\label{apdx:other}

In the main paper, we showed the performance of our method on the CIFAR-10 dataset. Here we show experiments on MNIST dataset.

For MNIST, a ResNet-18 encoder is first learned using the SimCLR \cite{SimCLR} objective for 100 epochs; then after applying data filtering and relabeling, a ResNet-18 classifier is trained using the SupCon \cite{SupCon} objective, on the cleansed dataset, for 100 epochs. Note that for MNIST, the pixels of the object (handwritten digits) are always white, with pixel value 255. WaNet, which will add some positive perturbation to already-white pixels (which are already at the maximum intensity), will not create an attacked image that is very different from the original image. So it will be hard for the WaNet attack to be successful on MNIST. Thus we did not include the WaNet attack in our MNIST experiment. For all the attacks, 1000 training samples are poisoned. 

Tab. \ref{tab:mnist} shows that for the additive attack and blended attack, our method decreases the ASR to close to 0 with only a slight drop in ACC. The performance of our method on the BadNet attack is not as good, but the ASR is still lower than 10\%.

\begin{table}[ht!]
	\scriptsize
	\begin{center}
			\begin{tabular}{ccccccc}
				\toprule
				& \multicolumn{2}{c}{additive} &
				\multicolumn{2}{c}{BadNet}& \multicolumn{2}{c}{blend} \\
				\cmidrule(lr){2-3} \cmidrule(lr){4-5}
				\cmidrule(lr){6-7} 
				&ASR&ACC&ASR&ACC&ASR&ACC
				\\\hline \hline
				None &91.9&99.50&99.9&99.95&100&99.95  \\
				kNN &0.2& 99.48&9.3&99.35&0.3&99.29 \\
				Energy &0&99.38&3.8&99.59&1.4&99.46  \\
		\bottomrule
		\end{tabular}
		\caption{Results on MNIST.}
		\label{tab:mnist}
	\end{center}
	\vspace{-0.12in}
\end{table}

\section{Robustness to adaptive attacks}
\label{apdx:adaptive}
Here we evaluate the robustness of our method under an adaptive attack scenario, where the attacker has access to the whole training set and full knowledge of our defense method. If an attacker can create backdoor images that are close to the target class clean images in the embedding space, those backdoor images can survive after our filtering method; then the adaptive attack might defeat our method. However, the attacker does not control the defense process (otherwise an attacker can always be successful). Moreover, given the above assumptions of the adaptive attack, the
attacker does not have access to the encoder trained using SimCLR loss in the first step of our defense method.  Thus, the attacker needs to train a ``surrogate'' of our encoder (We assume the attacker uses the same architecture as that used by the defender.). The attacker can learn an adaptive backdoor pattern by solving the following problem:

\begin{equation}\label{eq:adaptive}
    u^* = \argmin_{u}\frac{1}{|\mathcal{I} \backslash \mathcal{I}_t|} \sum_{i \in \mathcal{I} \backslash \mathcal{I}_t}||L(E(g(x_i, m, u))) - \bar{z}_t ||_2,
\end{equation}
where $g(\cdot)$ is the backdoor embedding function; here BadNet embedding is used, and $m$ is a mask fixed at top left corner of the image. $\mathcal{I}$ is the set of indices of all the images in the training set, with $\mathcal{I}_t$ the set of indices of images from the target class. $L(E(\cdot))$ is the ``surrogate'' encoder, and $\bar{z}_t$ is calculated as the average:

\begin{equation}
    \bar{z}_t = \frac{1}{|\mathcal{I}_t|} \sum_{j \in \mathcal{I}_t} L(E(x_j))
\end{equation}

Then by minimizing the objective function given in Eq. \ref{eq:adaptive}, the attacker can learn a backdoor pattern that aims to make the backdoor attacked images close to the target class images in the embedding space. 

\begin{table}[ht!]
	\scriptsize
	\begin{center}
			\begin{tabular}{ccccccc}
				\toprule
				& \multicolumn{2}{c}{None} &
				\multicolumn{2}{c}{kNN}& \multicolumn{2}{c}{Energy} \\
				\cmidrule(lr){2-3} \cmidrule(lr){4-5}
				\cmidrule(lr){6-7} 
				&ASR&ACC&ASR&ACC&ASR&ACC
				\\\hline \hline
				3x3 patch &96.1&93.08&9.1&91.66&7.1&92.01 \\
				7x7 patch &100& 93.80&11.8&91.59&32.3&91.92\\
				11 x 11 patch &99.9&94.49&33.4&91.86&37.9&91.82  \\
		\bottomrule
		\end{tabular}
		\caption{Adaptive attack results.}
		\label{tab:adaptive}
	\end{center}
	\vspace{-0.12in}
\end{table}

We created three poisoned CIFAR-10 datasets by varying the patch size of the BadNet pattern (3x3 patch, 7x7 patch, and 11x11 patch). The defense performance against adaptive attacks is shown in Tab. \ref{tab:adaptive}. The table shows that our method achieves robustness even under these adaptive attacks. 
%DJM-new -- very unclear what you mean here -- points 1) and 2)
% are unclear
%Since 1) it will be hard to learn a universal backdoor pattern that makes all the images close to the target class images in the embedding space. The learned backdoor pattern can make some attacked images close to the target class images, which makes more backdoor images survive from the filtering method and thus cause higher ASR than non-adaptive attack. 2) The pattern learned on the ``surrogate'' encoder is not transferable to the exact encoder we used to do filtering. Our method is hard to defeat even an attacker assumes full knowledge of our method. 

\section{Performance on clean dataset}
Tab. \ref{tab:clean_dataset} shows the performance comparison on a clean (unpoisoned) CIFAR-10 dataset. Our method has slightly lower ACC then DBD. Since our method will always reject some training examples even if the dataset is clean, it is reasonable in future work to first to detect if a dataset is in fact poisoned (using either existing or novel detection methods),
with the filtering method applied only when a poisoned dataset is detected.

\begin{table}[ht!]
	\scriptsize
	\begin{center}
			\begin{tabular}{ccccc}
				\toprule
				& None & DBD\cite{Decouple_yiming} & knn& energy\\\hline
				ACC & 94.45 &92.05 & 91.71 &91.74\\
		\bottomrule
		\end{tabular}
		\caption{Performance on clean CIFAR-10 dataset.}
		\label{tab:clean_dataset}
	\end{center}
	\vspace{-0.12in}
\end{table}

\section{Combining kNN and energy methods}
Tab. \ref{tab:performance} in the main paper shows the performance of our kNN based and energy based defense methods. Here we show the performance when combining the kNN and energy-based methods-- an image is accepted only when it is accepted by both the kNN and energy methods; an images is relabeled only when it is relabeled by both methods {\it and} when it is relabeled to the same class. Results in Tab. \ref{tab:knn_energy}, when compared to the results in the main paper, show that combining both methods does not lead to improved performance, which indicates that the samples accepted by both methods mostly overlap. 
\begin{table}[ht!]
	\scriptsize
	\begin{center}
			\begin{tabular}{cccccccc}
				\toprule
				 \multicolumn{2}{c}{Add} &
				\multicolumn{2}{c}{Patch}& \multicolumn{2}{c}{blend} & \multicolumn{2}{c}{WaNet}\\
				\cmidrule(lr){1-2} \cmidrule(lr){3-4}
				\cmidrule(lr){5-6} \cmidrule(lr){7-8}
	ASR&ACC&ASR&ACC&ASR&ACC&ASR&ACC
				\\\hline \hline
		3.3&91.23&3.5& 91.43 &4.2 & 91.29&2.2& 90.44		\\
		\bottomrule
		\end{tabular}
		\caption{Results of combining kNN and energy methods.}
		\label{tab:knn_energy}
	\end{center}
	\vspace{-0.12in}
\end{table}

\end{document}